\documentclass[11pt]{article}

\usepackage[preprint]{acl}

\usepackage{times}
\usepackage{latexsym}

\usepackage[T1]{fontenc}

\usepackage[utf8]{inputenc}

\usepackage{microtype}

\usepackage{inconsolata}

\usepackage{graphicx}
\usepackage{booktabs}
\usepackage{amsmath}
\usepackage{multirow}
\usepackage{caption}
\usepackage{subcaption}
\usepackage{amssymb}
\usepackage{adjustbox}
\usepackage[table]{xcolor}

\title{Learning the Cue or Learning the Word?\\ Analyzing Generalization in Metaphor Detection for Verbs} 

\author{
Sinan Kurtyigit$^{1, 2}$~~~Sabine Schulte im Walde$^{2}$~~~Alexander Fraser$^{1}$\\
$^{1}$School of Computation, Information and Technology, Technical University of Munich\\
$^{2}$Institute for Natural Language Processing, University of Stuttgart\\
\small
{\tt sinan.kurtyigit@tum.de}\;\;\;
\small
{\tt schulte@ims.uni-stuttgart.de}\;\;\;
\small
{\tt alexander.fraser@tum.de}
}

\begin{document}
\maketitle

\begin{abstract}
Metaphor detection models achieve strong benchmark performance, yet it remains unclear whether this reflects transferable generalization or lexical memorization. To address this, we analyze generalization in metaphor detection through RoBERTa, the shared backbone of many state-of-the-art systems, focusing on English verbs using the VU Amsterdam Metaphor Corpus. We introduce a controlled lexical hold-out setup where all instances of selected target lemmas are strictly excluded from fine-tuning, and compare predictions on these Held-out lemmas against Exposed lemmas (verbs seen during fine-tuning). While the model performs best on Exposed lemmas, it maintains robust performance on Held-out lemmas. Further analysis reveals that sentence context alone is sufficient to match full-model performance on Held-out lemmas, whereas static verb-level embeddings are not. Together, these results suggest that generalization is primarily driven by "learning the cue" (transferable contextual patterns), while "learning the word" (verb-specific memorization) provides an additive boost when lexical exposure is available.

\end{abstract}

\section{Introduction}
Metaphors allow abstract concepts to be expressed through concrete terms (cf. \textit{Conceptual Metaphor Theory} \citep{lakoff1980metaphors}), creating semantic tension between a verb and its arguments (e.g., \textit{grasp the meaning}). While state-of-the-art models \citep{deepmet, gofigure, melbert} achieve strong performance on benchmarks \citep{vua20}, it remains unclear whether this reflects genuine understanding or reliance on lexical memorization.

In this work, we analyze the generalization capabilities of a RoBERTa-based metaphor detection model on English verbs, targeting the backbone shared by many state-of-the-art systems \citep{deepmet, illinimet, melbert, roppt}. We specifically investigate whether the model generalizes by "learning the cue" (leveraging transferable contextual patterns), or merely relies on "learning the word" (recalling verb-specific associations observed during fine-tuning). This distinction is critical, as models relying on lexical exposure will struggle to handle the creative, novel expressions characteristic of real-world language.

Our contributions are threefold: (1) we introduce a controlled lexical hold-out setup on the VUA corpus \citep{vuac} to assess performance on Held-out lemmas entirely absent from fine-tuning; (2) we analyze whether sentence context alone or verb-level embeddings alone are sufficient to solve the task; and (3) we analyze the geometric structure of the embedding space to determine whether task-relevant organization extends to held-out lexical items.

Our experiments reveal that while performance is highest for Exposed lemmas, the model maintains robust performance on Held-out lemmas. This capability does not correlate with pre-training frequency, and Context-only experiments demonstrate that the model can detect metaphors even when the target verb is masked, whereas verb-level embeddings in isolation are insufficient. Geometric analysis further confirms that Held-out instances are projected into similar semantic subspaces as Exposed instances. Crucially, when verb identity is removed, performance on Exposed lemmas drops to match that of Held-out lemmas, suggesting a shared reliance on contextual cues while verb identity provides a supplementary boost.

We structure the remainder of this paper as follows: \S~\ref{sec:related_work} reviews related work in metaphor detection. \S~\ref{sec:setup} describes the experimental framework and the construction of the Exposed and Held-out splits. Our core analysis is presented in \S~\ref{sec:analysis}. Finally, \S~\ref{sec:discussion} explores the implications of our findings, followed by concluding remarks in \S~\ref{sec:conclusion}.

\section{Related Work}
\label{sec:related_work}

The VUA shared tasks \citep{vua18, vua20} provide the primary benchmark datasets and a reference point for the progress in token-level metaphor detection. Complementary datasets that are also widely used are TroFi \cite{trofi} and MOH-X \cite{mohx}.

From a modeling perspective, transformer-based architectures \citep{transformer} now dominate the field, particularly encoder-only models such as BERT \citep{bert} and its variants. A simple yet effective baseline is obtained by attaching a classification head to an encoder \citep{vua20, gofigure}. More sophisticated approaches incorporate additional information or modeling techniques. \citet{deepmet}, for example, frame metaphor detection as a reading comprehension problem, leveraging both global and local textual context as well as part-of-speech features. \citet{gofigure} enhance BERT with out-of-domain fine-tuning and auxiliary idiom detection, a task closely related to metaphorical language. Other approaches \citep{melbert, misnet, basicbert}, explicitly integrate linguistic metaphor identification theories such as \textit{Metaphor Identification Theory} \citep{mip} and \textit{Selectional Preference Violation } \citep{spv1975, spv1978}, to further guide detection.

More recently, large language models have been explored for metaphor detection, reflecting their broader adoption across NLP tasks. \citet{c4mmd} incorporate chain-of-thought prompting, while \citet{fuoli} investigate retrieval-augmented generation and prompt engineering in addition to classical fine-tuning strategies. \citet{liang} experiment with alternative prompting techniques to improve metaphor identification in LLMs.

\section{Experimental Setup}
\label{sec:setup}

\subsection{Task}
We define metaphor detection as a binary classification task at the word level. Given a sentence and a marked target word, the goal is to determine whether the target is used literally (0) or metaphorically (1) in context. For example, in \textit{"The committee absorbed the cost"}, the target \textit{absorbed} is metaphorical because it describes an abstract financial event using a physical process.

While applicable to all parts of speech, we focus exclusively on \textbf{verbs}, as they act as the structural core of sentences. Metaphors frequently arise from the semantic tension between a verb and its arguments, making verbs ideal for analyzing structural generalization. Following standard practice \citep{vua18, vua20}, we evaluate performance using the \textbf{binary F1 score} for the metaphorical class.

\subsection{Data}
We conduct our experiments on the VU Amsterdam Metaphor Corpus \citep[VUA]{vuac}, which serves as the standard benchmark for metaphor detection in the field. We rely on the VUA Verbs benchmark to establish model validity, and construct custom splits to assess generalization.

\subsubsection{VUA}
To ensure comparability with prior research, we adhere to the protocols of the VUA Shared Tasks \citep{vua18, vua20}, focusing on the \textbf{Verbs track}. We use the official gold labels to reconstruct the standard train and test splits, explicitly filtering the training set to remove any remaining non-verb samples to ensure strict alignment with our task definition.

Metaphoricity in the VUA corpus is annotated following the MIPVU procedure \citep{vuac}. A verb is labeled as metaphorical when its contextual meaning differs from a more basic or concrete sense, covering both salient figurative usages such as \textit{devour} in \textit{"She devoured the book"} and conventionalized expressions such as \textit{rise} in \textit{"Inflation has risen"}, where the figurative nature may no longer be consciously perceived by speakers.

\subsubsection{Lexical Hold-out}

\begin{table*}[!t]
    \centering
    \small
    \begin{tabular*}{\linewidth}{l@{\extracolsep{\fill}}rrrrl@{\extracolsep{\fill}}rrr}
        \toprule
        \multicolumn{4}{c}{\textbf{Exposed Set} ($n=10$ per lemma)} & & \multicolumn{4}{c}{\textbf{Held-out Set} ($n=20$ per lemma)} \\
        \cmidrule{1-4} \cmidrule{6-9}
        \textbf{Lemma} & \textbf{$N_{\text{VUA}}$} & \textbf{Met.\%$_{\text{VUA}}$} & \textbf{Met.\%$_{\text{Eval}}$} & & \textbf{Lemma} & \textbf{$N_{\text{VUA}}$} & \textbf{Met.\%$_{\text{VUA}}$} & \textbf{Met.\%$_{\text{Eval}}$} \\
        \midrule
        \multicolumn{4}{l}{\textit{Metaphorical-biased}} & & \multicolumn{4}{l}{\textit{Metaphorical-biased}} \\
        face       &  11 & 91\% & 90\% & & break      &  25 & 92\% & 90\% \\
        base       &  10 & 90\% & 90\% & & share      &  21 & 86\% & 85\% \\
        consider   &  13 & 85\% & 80\% & & feel       &  82 & 82\% & 80\% \\
        follow     &  13 & 85\% & 80\% & & lose       &  29 & 79\% & 80\% \\
        cover      &  14 & 79\% & 80\% & & reduce     &  38 & 79\% & 80\% \\
        set        &  12 & 75\% & 80\% & & regard     &  23 & 78\% & 80\% \\
        show       &  36 & 67\% & 70\% & & link       &  22 & 77\% & 75\% \\
        give       &  76 & 64\% & 60\% & & add        &  44 & 77\% & 75\% \\
        take       & 108 & 62\% & 60\% & & make       & 256 & 77\% & 75\% \\
        assume     &  13 & 62\% & 60\% & & rise       &  26 & 77\% & 75\% \\
        \midrule
        \multicolumn{4}{l}{\textit{Balanced}} & & \multicolumn{4}{l}{\textit{Balanced}} \\
        design     &  10 & 60\% & 60\% & & change     &  33 & 55\% & 55\% \\
        run        &  19 & 58\% & 60\% & & keep       &  70 & 54\% & 55\% \\
        leave      &  33 & 55\% & 50\% & & manage     &  32 & 53\% & 55\% \\
        cut        &  11 & 55\% & 50\% & & allow      &  42 & 52\% & 50\% \\
        fail       &  13 & 54\% & 50\% & & hold       &  69 & 52\% & 50\% \\
        see        &  99 & 45\% & 50\% & & find       & 130 & 49\% & 50\% \\
        move       &  18 & 44\% & 40\% & & bring      &  73 & 48\% & 50\% \\
        turn       &  35 & 43\% & 40\% & & put        & 109 & 48\% & 50\% \\
        open       &  17 & 41\% & 40\% & & offer      &  23 & 43\% & 45\% \\
        come       & 121 & 40\% & 40\% & & produce    &  37 & 43\% & 45\% \\
        \midrule
        \multicolumn{4}{l}{\textit{Literal-biased}} & & \multicolumn{4}{l}{\textit{Literal-biased}} \\
        meet       &  10 & 40\% & 40\% & & play       &  45 & 24\% & 25\% \\
        decide     &  11 & 36\% & 40\% & & carry      &  43 & 23\% & 25\% \\
        include    &  15 & 33\% & 30\% & & call       &  65 & 23\% & 25\% \\
        work       &  37 & 32\% & 30\% & & go         & 660 & 22\% & 20\% \\
        involve    &  12 & 25\% & 20\% & & identify   &  25 & 20\% & 20\% \\
        achieve    &  10 & 20\% & 20\% & & look       & 196 & 20\% & 20\% \\
        mean       &  91 & 16\% & 20\% & & send       &  21 & 19\% & 20\% \\
        talk       &  26 & 15\% & 20\% & & stay       &  38 & 18\% & 20\% \\
        write      &  10 & 10\% & 10\% & & collect    &  28 & 14\% & 15\% \\
        try        &  32 &  9\% & 10\% & & expect     &  53 &  8\% & 10\% \\
        \bottomrule
    \end{tabular*}
    \caption{Target lemmas selected for the Exposed (left) and Held-out (right) evaluation sets. $N_{\text{VUA}}$ indicates the original instance count in the VUA Verbs train split. Met.\%$_{\text{VUA}}$ and Met.\%$_{\text{Eval}}$ report the percentage of metaphorical instances before and after stratified downsampling, respectively.}
    \label{tab:target_lemmas}
\end{table*}

To systematically evaluate generalization capabilities, we derive two parallel diagnostic sets from the VUA Verbs datasets: a \textbf{Held-out} set (target lemmas excluded from fine-tuning) and an \textbf{Exposed} set (target lemmas included in fine-tuning). We explicitly distinguish between the \textbf{verb lemma} (the lexical type) and the \textbf{verb instance} (the specific usage in context). Metaphoricity is a property of the instance and is evaluated at that level, while the split into Exposed and Held-out sets is defined at the lemma level, such that all instances targeting a given lemma are either consistently excluded from or included in fine-tuning.

\paragraph{Held-out Lemmas}
We select 30 target lemmas to be completely excluded from fine-tuning. Candidates are drawn from the training split (min. frequency 20) and stratified into three categories: 10 predominantly metaphorical, 10 predominantly literal, and 10 balanced. The selection process is systematic: we first rank all lemmas by their metaphoricity ratio in the training set (see Table \ref{tab:target_lemmas}). The 10 highest-ranked lemmas form the \textit{Metaphorical-biased} subset; the \textit{Literal-biased} subset mirrors its inverse distribution (e.g., a lemma with 90\% metaphorical usage is paired with one at 10\%); and the \textit{Balanced} subset comprises the 10 lemmas closest to a 50/50 split.\footnote{Due to limited data size, lemma selection is deterministic.} All instances targeting these lemmas are removed from the training data to create the \textbf{Filtered Train Set}. The withheld instances form the \textbf{Held-out Evaluation Set}, serving as a strict benchmark for generalization to novel verbs.

\paragraph{Exposed Lemmas} 
We select a parallel control set of 30 lemmas following the same systematic ranking approach. To measure the benefit of lexical exposure without testing on memorized instances, these lemmas appear in the training split (min. frequency 10),\footnote{The threshold is lowered from 20 to 10 due to insufficient candidates in the training-test vocabulary intersection.} but are evaluated on distinct instances from the test split. All training instances remain in the Filtered Train Set, while distinct test instances form the \textbf{Exposed Evaluation Set}.

\paragraph{Evaluation Set Downsampling}
To prevent frequent lemmas such as \textit{go} from dominating the results, we downsample the evaluation data. We draw 20 instances per lemma for the Held-out set (Total: 600) and 10 instances per lemma for the Exposed set (Total: 300). Downsampling is stratified to preserve the original class distribution of each lemma.\footnote{Downsampling is the only stochastic step in dataset construction and is seeded for reproducibility.}

\paragraph{Selected Lemmas}
Table \ref{tab:target_lemmas} presents the selected target lemmas and their distribution statistics. For the Held-out set, the categories are distinctly stratified: metaphorical-biased verbs range from 92\% to 77\% metaphoricity, balanced verbs cluster tightly between 55\% and 43\%, and literal-biased verbs range from 24\% down to 8\%.

In the Exposed set, however, data scarcity within the intersection of training and test vocabularies necessitates looser thresholds. Consequently, the category boundaries are less distinct. Metaphorical-biased verbs extend down to 62\%, and balanced verbs span from 60\% down to 40\%. This creates a slight overlap with the literal-biased category, which begins at 40\% (e.g., meet) and extends down to 9\%. For instance, \textit{come} falls into the balanced category with 40\% metaphoricity, matching the upper bound of the literal category.

\subsubsection{Dataset Statistics}
Table \ref{tab:data-stats} summarizes the dataset statistics. The standard VUA Verbs datasets are large but imbalanced (approx. 28--30\% metaphorical). Our filtering process removes 2,358 samples (13\%) from the training data to create the Filtered Train Set (14,882 samples), retaining sufficient data for learning the task.

\begin{table}[]
\centering
\begin{tabular*}{\linewidth}{l@{\extracolsep{\fill}}rrr}
\toprule
\textbf{Dataset Split} & \textbf{$N_{\text{Samples}}$} & \textbf{Met. \%} & \textbf{$N_{\text{Lemmas}}$} \\
\midrule
\textit{Standard VUA} & & & \\
Train & 17240 & 28\% & 2190 \\
Test & 5873 & 30\% & 1111 \\
\midrule
\textit{Lexical Hold-out} & & & \\
Filtered Train & 14882 & 26\% & 2163 \\
Exposed Eval. & 300 & 49\% & 30 \\
Held-out Eval. & 600 & 50\% & 30 \\
\bottomrule
\end{tabular*}
    \caption{Statistics for the VUA Verbs datasets, reporting the number of samples, proportion of metaphorical labels, and count of unique target lemmas. The top section displays the standard benchmark splits, while the bottom section displays the constructed lexical hold-out splits.}
\label{tab:data-stats}
\end{table}

In contrast, our diagnostic sets are strictly balanced. Both the Held-out (600 samples) and Exposed (300 samples) evaluation sets approximate a 50/50 class distribution. This balance establishes a clear \textbf{Random Baseline}: a model predicting uniformly at random yields a binary F1 of approximately .500. We use this threshold to distinguish genuine signal from chance during the analysis.

\subsection{Model Architecture}
\label{subsec:model}
We use a pre-trained RoBERTa \citep{roberta} encoder paired with a single linear classification head. RoBERTa is widely used as the shared backbone across metaphor detection systems \citep{deepmet, illinimet, melbert, roppt}, many of which extend it with auxiliary tasks, external knowledge, or specialized interaction layers. Analyzing the generalization behavior of the backbone itself is therefore directly relevant to the field, and keeping the architecture minimal avoids confounding factors, enabling a transparent and interpretable analysis.

\begin{figure}
\centering
\includegraphics[width=1\linewidth]{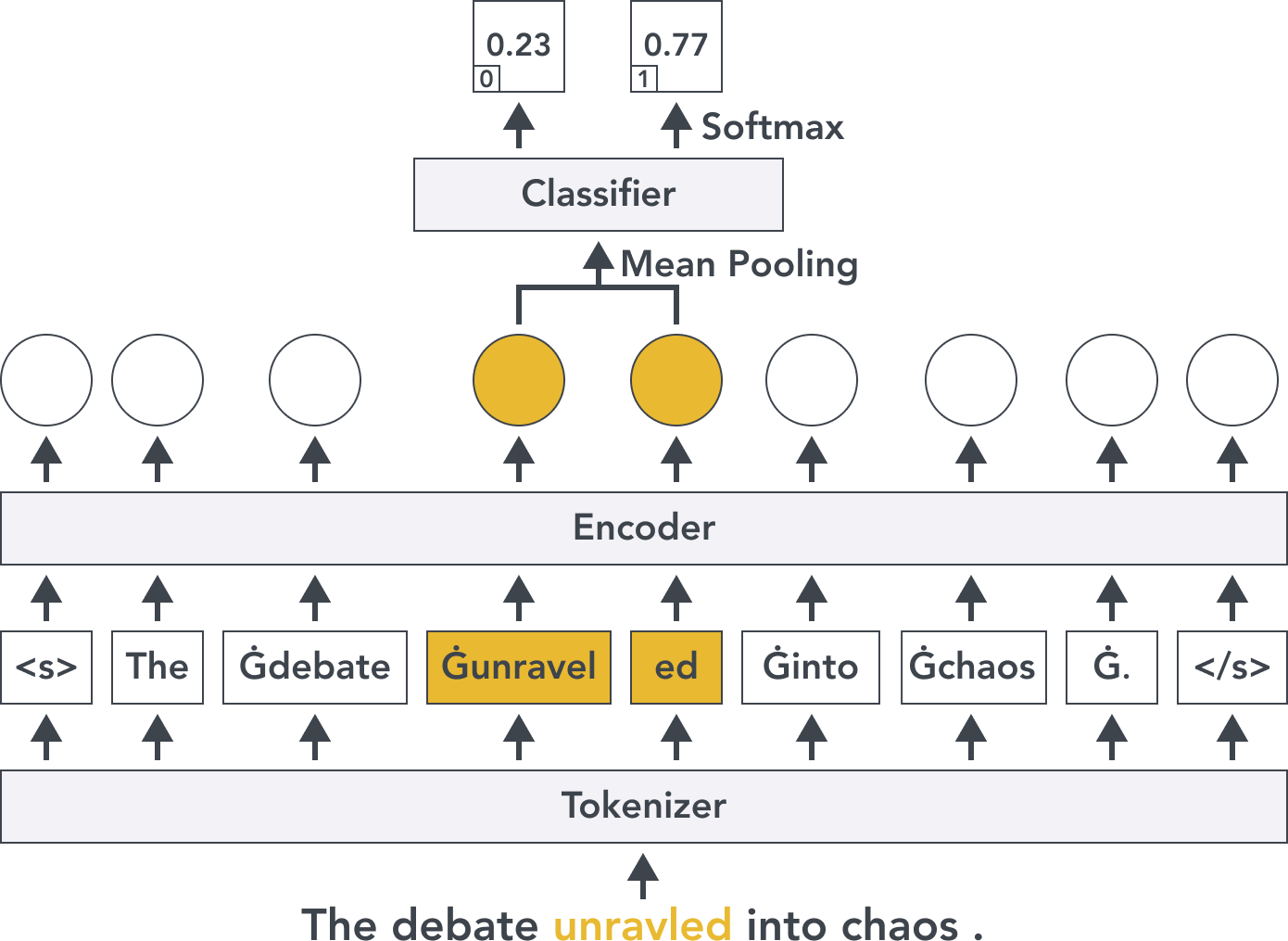}
\caption{Model architecture. The input sentence is tokenized and passed through the RoBERTa encoder. Hidden states corresponding to the target verb's sub-word tokens (e.g., \textit{unravel} + \textit{ed}) are aggregated via mean pooling into a \textbf{Contextualized Target Representation}, which is passed to a linear classifier to predict metaphoricity.}
\label{fig:model-architecture}
\end{figure}

As illustrated in Figure \ref{fig:model-architecture}, given an input sentence such as \textit{"The debate unraveled into chaos"}, the text is tokenized into sub-word units, where the target verb may be decomposed into multiple tokens (e.g., \textit{unravel} and \textit{ed}). These tokens and their surrounding context are passed through the encoder, generating a sequence of contextualized hidden states.

To derive a fixed-size input for the classifier, we apply \textbf{mean pooling} over the hidden states corresponding to the target verb's sub-word tokens, producing the \textbf{Contextualized Target Representation}. This representation encapsulates the target verb's meaning as shaped by its surrounding context. It is passed as the sole input to the linear classification layer, which produces logits normalized via Softmax into final probabilities for the literal (0) and metaphorical (1) classes.

\subsection{Fine-Tuning}
Our fine-tuning procedure consists of two distinct phases: first, we establish model stability and select a representative configuration using the Standard VUA benchmark; and second, we apply this configuration to the Filtered train set for analysis.

\paragraph{Stability and Seed Selection} 
We perform a modest hyperparameter search on the standard VUA Verbs train set to identify a reliable configuration. For all training runs, we reserve 10\% of the training data as a validation split and apply a weight of 3.0 to the metaphorical class to address label imbalance. Using the obtained set of parameters (5 epochs, batch size 32, learning rate 4e-5, weight decay 0.02, linear scheduler with 0.1 warmup), we fine-tune 10 models from scratch, each initialized with a different random seed. For each run, we retain the checkpoint that achieves the highest F1 score on the validation split.

Table \ref{tab:seed-stability} summarizes the model's performance across ten independent runs on the validation set. We observe a narrow performance range, with F1 scores between .706 and .734 ($\mu = .717,\ \sigma = .009$). This minimal variance confirms that the model architecture and fine-tuning process are robust to random initialization. To ensure that our subsequent analysis reflects representative behavior rather than a performance outlier, we select the median-performing run. As seeds 42 and 5555 are tied for the median F1 score, we select \textbf{seed 42} as the representative model for our study based on its superior validation performance.

\begin{table}[]
    \centering
    \begin{tabular*}{\linewidth}{r@{\extracolsep{\fill}}ccc}
        \toprule
        \textbf{Seed} & \textbf{Prec.} & \textbf{Rec.} & \textbf{F1} \\
        \midrule
        9999 & .738 & .676 & .706 \\
        314 & .760 & .667 & .710 \\
        999 & .736 & .689 & .712 \\
        7 & .706 & .721 & .713 \\
        5555 & .674 & .763 & .716 \\
        \textbf{42} & \textbf{.681} & \textbf{.755} & \textbf{.716} \\
        8765 & .740 & .696 & .717 \\
        2025 & .750 & .692 & .719 \\
        1234 & .743 & .718 & .731 \\
        777 & .719 & .749 & .734 \\
        \midrule
        \textbf{Mean} & .725 & .712 & .717 \\
        \textbf{Std} & .029 & .034 & .009 \\
        \bottomrule
    \end{tabular*}
    \caption{Model performance across 10 random seeds on the validation split. The table reports Precision, Recall, and F1 scores for each run, sorted by F1. Summary statistics (Mean, Std) are provided at the bottom, and the selected median seed (42) is highlighted.}
    \label{tab:seed-stability}
\end{table}

\paragraph{Application to Hold-Out Experiments} For the generalization experiments described in \S\ref{sec:analysis}, we re-initialize the model using the selected \textbf{seed 42} and fine-tune it from scratch on the \textbf{Filtered Train Set}, following the same validation protocol to select the best checkpoint. This consistency ensures that any performance drops we observe are due to the data filtering (the hold-out), not due to switching to a less stable random seed.

\subsection{Validation on Standard Benchmarks}
Before analyzing generalization, we verify the validity of our framework.

First, we evaluate our representative model on the official VUA Verbs test set (Table \ref{tab:baseline_validation}). Our F1 score of .716 is lower than systems that build on top of the RoBERTa backbone with additional components, such as DeepMet (.804) \citep{deepmet} and GoFigure (.775) \citep{gofigure}, but remains competitive within the Shared Task context \citep{vua20}. Since our goal is to analyze the backbone itself rather than to maximize performance, we consider this sufficient to serve as a valid testbed.

\begin{table}[]
    \centering
    \begin{tabular*}{\linewidth}{l@{\extracolsep{\fill}}ccc}
        \toprule
        \textbf{Model} & \textbf{Prec.} & \textbf{Rec.} & \textbf{F1} \\
        \midrule
        \textit{Shared Task Baselines} & & & \\
        DeepMet & .789 & .819 & .804 \\
        Go Figure! & .732 & .823 & .775 \\
        \midrule
        \textit{Our Models} & & & \\
        Ours (Standard) & .681 & .755 & .716 \\
        Ours (Filtered) & .754 & .663 & .706 \\
        \bottomrule
    \end{tabular*}
    \caption{Performance on the VUA Verbs Test Set reporting Precision, Recall, and F1. The top section lists results from the 2020 VUA Shared Task. The bottom section displays our model trained on the full dataset (Standard) and the filtered train set (filtered).}

    \label{tab:baseline_validation}
\end{table}

Second, we assess the impact of removing the 30 Held-out lemmas, or more specifically, the instances targeting them, from the training data. As shown in Table \ref{tab:baseline_validation}, the model trained on the \textbf{Filtered Train Set} maintains performance nearly identical to the full model ($\Delta$ .010). This confirms that despite reduced lexical exposure, the model retains the necessary predictive capability for our generalization experiments.

\section{Analysis of Generalization Mechanism}
\label{sec:analysis}

Having established the experimental setup, we use the Exposed and Held-out evaluation sets to analyze the extent and nature of the model’s generalization behavior. Our goal is to determine whether the model exhibits generalization capabilities to novel lexical items and to investigate whether this performance reflects the acquisition of transferable contextual cues ("learning the cue") or a reliance on verb-specific associations learned through lexical exposure ("learning the word").

The analysis proceeds in three steps. First, we compare performance on Exposed and Held-out verbs and examine whether performance on Held-out verbs can be explained by pre-training frequency (\S~\ref{subsec:gen_performance}). Second, we isolate sentence context and verb identity to test whether either source of information is sufficient on its own for metaphor detection (\S~\ref{subsec:signal_sources}). Finally, we analyze the geometry of the learned representation space to assess whether task-relevant structure is present in the Contextualized Target Representations, and whether this structure extends to Held-out verbs (\S~\ref{subsec:geometry}).

\subsection{Generalization Performance and Pre-training Frequency}
\label{subsec:gen_performance}

We begin by comparing model performance on the Held-out set against performance on the Exposed set and the Random Baseline established in \S~\ref{sec:setup}. The Exposed set contains verbs that were included during fine-tuning and therefore reflects performance when the model has direct lexical supervision. The Random Baseline (.500) represents performance in the absence of any task-relevant signal.

\begin{table}[]
\centering
\begin{tabular*}{\linewidth}{l@{\extracolsep{\fill}}ccc}
\toprule
\textbf{Evaluation Set} & \textbf{Prec.} & \textbf{Rec.} & \textbf{F1} \\
\midrule
\textit{Full Model} \\
Exposed & .864 & .776 & .817 \\
Held-out & .811 & .573 & .672 \\
\midrule
\textit{Context-only} \\
Exposed & .743 & .748 & .746 \\
Held-out & .714 & .650 & .681 \\
\midrule
\textit{Word-only} \\
Exposed & .586 & .837 & .689 \\
Held-out & .533 & .570 & .551 \\
\midrule
Random & .500 & .500 & .500 \\
\bottomrule
\end{tabular*}
\caption{Performance on the Exposed and Held-out evaluation sets, reporting Precision, Recall, and F1. The table is divided by model variant to isolate information sources: the Full Model (top) serves as the primary benchmark; Context-only (middle) assesses the sufficiency of structural cues; and Word-only (bottom) evaluates the predictive power of lexical identity alone. The bottom row presents the Random Baseline.}
\label{tab:generalization_performance}
\end{table}

On the Exposed set, the model achieves an F1 score of .817, indicating strong performance when the target verbs have been observed during fine-tuning. Performance on the Held-out set decreases to .672 but remains well above the random baseline. This shows that the model retains access to task-relevant signal even when verb-specific supervision is unavailable. At the same time, the gap between Exposed and Held-out performance highlights the clear benefit of "learning the word" during fine-tuning.

While these results suggest that Held-out performance cannot be explained by fine-tuning memorization alone, they do not rule out the possibility that it is driven by lexical knowledge acquired during large-scale pre-training. To investigate this, we analyze the relationship between verb frequency and model performance, using frequency as a proxy for the strength of pre-training priors. We assume that verbs encountered more frequently during pre-training are associated with higher-quality representations. As exact pre-training statistics are unavailable, we rely on frequency estimates from the wordfreq library \citep{wordfreq}, which aggregates counts across multiple large-scale corpora.
    
We compute the Spearman correlation between per-lemma F1 scores and frequency estimates. For verbs in the Exposed set, we observe a moderate positive correlation ($\rho=.420,\ p=.021$), consistent with the intuition that more frequent verbs are easier to fine-tune. For verbs in the Held-out set, however, the correlation is negligible and statistically non-significant ($\rho = -.127,\ p = .504$). This lack of correlation indicates that performance on novel verbs does not systematically increase with pre-training frequency, suggesting that generalization is not driven by frequency bias in the pre-trained encoder.

Taken together, these findings indicate that the model’s performance on Held-Out verbs goes beyond memorized verb-specific information acquired during fine-tuning and does not simply scale with pre-training frequency.

\subsection{Sufficiency of Context and Verb Identity}
\label{subsec:signal_sources}

Since verb frequency does not appear to account for performance on the Held-out set, we next examine which information sources are sufficient to support prediction. We focus on two questions: (1) whether sentence context alone can support metaphor detection ("learning the cue"), and (2) whether verb-level representations encode a reusable signal independent of context.

We compare the full fine-tuned model against two controlled variants:
\begin{enumerate}
    \item \textbf{Context-only:} We evaluate the fine-tuned model with the target verb replaced by a \texttt{<mask>} token. This removes access to verb identity, allowing us to test whether contextual cues alone are sufficient.
    \item \textbf{Word-only:} We train a logistic regression classifier on static embeddings extracted from the token lookup layer (Layer 0) of the fine-tuned model. Unlike the Contextualized Target Representation used by the full model, these embeddings are not passed through the encoder and thus contain no contextual information. This tests whether verb identity alone is predictive, independent of context.
\end{enumerate}

Table \ref{tab:generalization_performance} reports F1 scores under these conditions. On the Exposed set, performance decreases stepwise from the Full model (.817) to Context-only (.746) to Word-only (.689). This demonstrates that when lexical exposure is available, both contextual information and verb-level representations are sufficient to support reasonably strong performance.

For Held-out verbs, the pattern changes drastically. The Context-only model (.681) matches the performance of the full model (.671), showing that contextual information alone is sufficient to recover predictive performance in the absence of verb-specific supervision. In contrast, the Word-only variant drops to .551, close to chance level. This indicates that while "learning the word" aids exposed verbs, it does not provide a sufficient signal for generalization. Instead, the model's ability to handle Held-out verbs appears driven primarily by transferable contextual cues.

Overall, these results show a clear asymmetry between the two information sources: verb-level representations support prediction only for Exposed verbs, whereas contextual cues generalize robustly to Held-out verbs and are sufficient to drive model performance when verb identity is unavailable.

\subsection{Geometric Structure of the Representation Space}
\label{subsec:geometry}

Finally, we investigate whether the model's reliance on context is reflected in the geometry of its internal Contextualized Target Representations, asking whether task-relevant organization exists independently of the learned classifier weights. If the model encodes transferable cues, instances with the same label should occupy similar regions of the representation space, even for verbs withheld during fine-tuning.

To address this, we pass each training sample through the fine-tuned model and extract its Contextualized Target Representation (as described in \S~\ref{subsec:model}), forming a reference space. Evaluation samples are embedded in the exact same way to allow direct geometric comparison. For each evaluation instance, we retrieve its $k=10$ nearest neighbors using cosine similarity and assess the representational structure with two measures: (1) \textit{Neighborhood Purity}, the proportion of neighbors sharing the same label as the target instance; and (2) a \textit{$k$-Nearest Neighbor ($k$-NN) classifier}, probing whether label information is recoverable without a learned decision boundary. We mirror the conditions in \S~\ref{subsec:signal_sources} by comparing the \textbf{Full} model and a \textbf{Context-only} variant.

Table~\ref{tab:geometry} reports the results. For the Full model, purity is high for Exposed verbs (.789) and remains substantial for Held-out verbs (.690). This mirrors the trends in \S~\ref{subsec:gen_performance}: although lexical exposure provides a clear advantage, Held-out instances are still embedded near training examples with the same label, indicating that the model successfully projects novel verbs into task-relevant neighborhoods.

\begin{table}[]
\centering
\begin{tabular*}{\linewidth}{l@{\extracolsep{\fill}}rr}
\toprule
\textbf{Evaluation Set} & \textbf{Purity} & \textbf{$k$-NN F1} \\
\midrule
\textit{Full Model}\\
Exposed & .789 & .788 \\
Held-out & .690 & .637 \\
\midrule
\textit{Context-only}\\
Exposed & .638 & .505 \\
Held-out & .624 & .544 \\
\bottomrule
\end{tabular*}
\caption{Geometric analysis of the Contextualized Target Representations using a $k$-Nearest Neighbor probe ($k=10$). We report Neighborhood Purity and $k$-NN F1 score for Exposed and Held-out verbs, comparing the Full Model (top) against the Context-only variant (bottom) to assess the impact of verb identity on the embedding space structure.}
\label{tab:geometry}
\end{table}

Crucially, in the Context-only condition, purity scores for Exposed (.638) and Held-out (.624) verbs converge, suggesting that a transferable geometric organization exists independently of verb identity.

The $k$-NN classification results reveal a related distinction. For the Full model, the probe achieves F1 scores of .788 on Exposed verbs and .637 on Held-out verbs, closely tracking the trained classifier's performance and indicating that the embedding geometry captures the essential decision boundary. In contrast, performance drops sharply for the Context-only variant (.505 Exposed; .544 Held-out), suggesting that while context is sufficient for local label consistency (purity), verb identity is required to sharpen the global separation of the space. Without this lexical signal, the representations are less distinctly partitioned, requiring the learned linear classifier to effectively extract the task-relevant signal.

Overall, this geometric analysis suggests that contextual cues provide a shared structural baseline for generalization, while verb identity acts as a complementary signal that strengthens class separability.

\section{Discussion}
\label{sec:discussion}
Across all three analyses, a consistent pattern emerges. Performance is highest on the Exposed set, lower but still noticeably above the random baseline on the Held-out set. This hierarchy appears not only in classification metrics but also in representation-level measures such as neighborhood purity and $k$-NN performance. This suggests that while the model benefits from "learning the word" (lexical exposure), it successfully generalizes by "learning the cue" (transferable contextual patterns), preventing performance from collapsing to chance on novel verbs.
The Context-only (masked) condition provides the key to disentangling these factors. By masking the target verb, we force the model to rely solely on "the cue," removing access to "the word." Under this condition, results across all metrics converge: performance on Exposed verbs drops to match that of Held-out verbs, and the gap largely disappears. This convergence is observed consistently for classification performance, embedding purity, and $k$-NN probing.
This "downward convergence" indicates that the superior performance on Exposed verbs is driven by the additive benefit of verb identity, rather than fundamentally different contextual processing. When verb identity is unavailable, the model processes Exposed and Held-out verbs similarly, relying on the same shared contextual cues.
Taken together, these results characterize the model as a hybrid system. It builds a robust floor of generalization through contextual representations ("learning the cue"), which allows it to handle unseen verbs. Simultaneously, it exploits specific lexical associations ("learning the word") to boost performance when the target verb has been observed during fine-tuning. The masking analysis demonstrates that while verb identity is a powerful performance amplifier, it is not a strict prerequisite for metaphor detection.

\section{Conclusion}
\label{sec:conclusion}

We presented a systematic analysis of the generalization capabilities of a RoBERTa-based metaphor detection model on verbs not observed during fine-tuning. Using a controlled lexical hold-out setup on the VU Amsterdam Metaphor Corpus, we demonstrated that while lexical exposure consistently improves performance, the model maintains robust performance on Held-out verbs through contextual information alone. This pattern holds across multiple analyses, including classification performance and representation-level probes.

Our results indicate that generalization cannot be explained solely by fine-tuning memorization, nor does it correlate with verb frequency as a proxy for pre-training exposure. Crucially, when verb identity is masked, performance on Exposed verbs drops to the level of Held-out verbs, causing the performance gap to disappear. This convergence suggests that the underlying contextual signal is shared across both conditions, while verb identity serves as an additive, rather than essential, source of information.

Geometric analysis supports this interpretation. Contextualized Target Representations of Held-out verbs exhibit task-relevant structure similar to that of Exposed verbs, confirming that the encoder projects novel metaphors into meaningful semantic subspaces.

Overall, our findings characterize the model as a hybrid system that generalizes by "learning the cue" (relying on transferable contextual patterns to handle novel verbs) while simultaneously "learning the word" to refine predictions when specific lexical associations are available.

\clearpage

\section*{Limitations}
Our study has several limitations. First, we rely on a single, standard encoder-based architecture; it remains unclear whether our findings regarding structural generalization extend to more complex state-of-the-art approaches or the decoder-only models increasingly dominant in NLP. Second, our analysis is restricted to English verbs. Future work should expand to other languages and parts of speech, ideally utilizing datasets with more fine-grained control over verb-object relations to enable a more robust analysis of selectional preferences. Finally, while our masking experiments demonstrate that context is sufficient for performance, we do not decompose the contextual signal to determine exactly which features (and to what extent) drive the model's predictions.



\bibliography{custom}

@book{vuac,
author = {Steen, Gerard and Dorst, Lettie and Herrmann, J. and Kaal, Anna and Krennmayr, Tina and Pasma, Trijntje},
year = {2010},
month = {06},
pages = {},
title = {A method for linguistic metaphor identification: From MIP to MIPVU},
isbn = {9789027239037},
doi = {10.1075/celcr.14}
}

@inproceedings{vua18,
    title = "A Report on the 2018 {VUA} Metaphor Detection Shared Task",
    author = "Leong, Chee Wee (Ben)  and
      Beigman Klebanov, Beata  and
      Shutova, Ekaterina",
    editor = "Beigman Klebanov, Beata  and
      Shutova, Ekaterina  and
      Lichtenstein, Patricia  and
      Muresan, Smaranda  and
      Wee, Chee",
    booktitle = "Proceedings of the Workshop on Figurative Language Processing",
    month = jun,
    year = "2018",
    address = "New Orleans, Louisiana",
    publisher = "Association for Computational Linguistics",
    url = "https://aclanthology.org/W18-0907/",
    doi = "10.18653/v1/W18-0907",
    pages = "56--66"
}

@inproceedings{vua20,
    title = "A Report on the 2020 {VUA} and {TOEFL} Metaphor Detection Shared Task",
    author = "Leong, Chee Wee (Ben)  and
      Beigman Klebanov, Beata  and
      Hamill, Chris  and
      Stemle, Egon  and
      Ubale, Rutuja  and
      Chen, Xianyang",
    editor = "Klebanov, Beata Beigman  and
      Shutova, Ekaterina  and
      Lichtenstein, Patricia  and
      Muresan, Smaranda  and
      Wee, Chee  and
      Feldman, Anna  and
      Ghosh, Debanjan",
    booktitle = "Proceedings of the Second Workshop on Figurative Language Processing",
    month = jul,
    year = "2020",
    address = "Online",
    publisher = "Association for Computational Linguistics",
    url = "https://aclanthology.org/2020.figlang-1.3/",
    doi = "10.18653/v1/2020.figlang-1.3",
    pages = "18--29"
}

@inproceedings{deepmet,
    title = "{D}eep{M}et: A Reading Comprehension Paradigm for Token-level Metaphor Detection",
    author = "Su, Chuandong  and
      Fukumoto, Fumiyo  and
      Huang, Xiaoxi  and
      Li, Jiyi  and
      Wang, Rongbo  and
      Chen, Zhiqun",
    editor = "Klebanov, Beata Beigman  and
      Shutova, Ekaterina  and
      Lichtenstein, Patricia  and
      Muresan, Smaranda  and
      Wee, Chee  and
      Feldman, Anna  and
      Ghosh, Debanjan",
    booktitle = "Proceedings of the Second Workshop on Figurative Language Processing",
    month = jul,
    year = "2020",
    address = "Online",
    publisher = "Association for Computational Linguistics",
    url = "https://aclanthology.org/2020.figlang-1.4/",
    doi = "10.18653/v1/2020.figlang-1.4",
    pages = "30--39"
}

@inproceedings{gofigure,
    title = "Go Figure! Multi-task transformer-based architecture for metaphor detection using idioms: {ETS} team in 2020 metaphor shared task",
    author = "Chen, Xianyang  and
      Leong, Chee Wee (Ben)  and
      Flor, Michael  and
      Beigman Klebanov, Beata",
    editor = "Klebanov, Beata Beigman  and
      Shutova, Ekaterina  and
      Lichtenstein, Patricia  and
      Muresan, Smaranda  and
      Wee, Chee  and
      Feldman, Anna  and
      Ghosh, Debanjan",
    booktitle = "Proceedings of the Second Workshop on Figurative Language Processing",
    month = jul,
    year = "2020",
    address = "Online",
    publisher = "Association for Computational Linguistics",
    url = "https://aclanthology.org/2020.figlang-1.32/",
    doi = "10.18653/v1/2020.figlang-1.32",
    pages = "235--243"
}

@inproceedings{melbert,
    title = "{M}el{BERT}: Metaphor Detection via Contextualized Late Interaction using Metaphorical Identification Theories",
    author = "Choi, Minjin  and
      Lee, Sunkyung  and
      Choi, Eunseong  and
      Park, Heesoo  and
      Lee, Junhyuk  and
      Lee, Dongwon  and
      Lee, Jongwuk",
    editor = "Toutanova, Kristina  and
      Rumshisky, Anna  and
      Zettlemoyer, Luke  and
      Hakkani-Tur, Dilek  and
      Beltagy, Iz  and
      Bethard, Steven  and
      Cotterell, Ryan  and
      Chakraborty, Tanmoy  and
      Zhou, Yichao",
    booktitle = "Proceedings of the 2021 Conference of the North American Chapter of the Association for Computational Linguistics: Human Language Technologies",
    month = jun,
    year = "2021",
    address = "Online",
    publisher = "Association for Computational Linguistics",
    url = "https://aclanthology.org/2021.naacl-main.141/",
    doi = "10.18653/v1/2021.naacl-main.141",
    pages = "1763--1773"
}

@article{mip,
author = {Crisp, Peter and Gibbs, Raymond and Deignan, Alice and Low, Graham and Steen, Gerard and Cameron, Lynne and Semino, Elena and Grady, Joe and Cienki, Alan and Kövecses, Zoltán and Group, The},
year = {2007},
month = {01},
pages = {},
title = {MIP: A method for identifying metaphorically used words in discourse},
volume = {22},
journal = {Metaphor and Symbol},
doi = {10.1207/s15327868ms2201_1}
}

@inproceedings{basicbert,
    title = "Metaphor Detection via Explicit Basic Meanings Modelling",
    author = "Li, Yucheng  and
      Wang, Shun  and
      Lin, Chenghua  and
      Guerin, Frank",
    editor = "Rogers, Anna  and
      Boyd-Graber, Jordan  and
      Okazaki, Naoaki",
    booktitle = "Proceedings of the 61st Annual Meeting of the Association for Computational Linguistics (Volume 2: Short Papers)",
    month = jul,
    year = "2023",
    address = "Toronto, Canada",
    publisher = "Association for Computational Linguistics",
    url = "https://aclanthology.org/2023.acl-short.9/",
    doi = "10.18653/v1/2023.acl-short.9",
    pages = "91--100"
}

@inproceedings{misnet,
    title = "Metaphor Detection via Linguistics Enhanced {S}iamese Network",
    author = "Zhang, Shenglong  and
      Liu, Ying",
    editor = "Calzolari, Nicoletta  and
      Huang, Chu-Ren  and
      Kim, Hansaem  and
      Pustejovsky, James  and
      Wanner, Leo  and
      Choi, Key-Sun  and
      Ryu, Pum-Mo  and
      Chen, Hsin-Hsi  and
      Donatelli, Lucia  and
      Ji, Heng  and
      Kurohashi, Sadao  and
      Paggio, Patrizia  and
      Xue, Nianwen  and
      Kim, Seokhwan  and
      Hahm, Younggyun  and
      He, Zhong  and
      Lee, Tony Kyungil  and
      Santus, Enrico  and
      Bond, Francis  and
      Na, Seung-Hoon",
    booktitle = "Proceedings of the 29th International Conference on Computational Linguistics",
    month = oct,
    year = "2022",
    address = "Gyeongju, Republic of Korea",
    publisher = "International Committee on Computational Linguistics",
    url = "https://aclanthology.org/2022.coling-1.364/",
    pages = "4149--4159"
}

@article{spv1978,
title = {Making preferences more active},
journal = {Artificial Intelligence},
volume = {11},
number = {3},
pages = {197-223},
year = {1978},
issn = {0004-3702},
doi = {https://doi.org/10.1016/0004-3702(78)90001-2},
url = {https://www.sciencedirect.com/science/article/pii/0004370278900012},
author = {Yorick Wilks}
}

@inproceedings{c4mmd,
    title = "Exploring Chain-of-Thought for Multi-modal Metaphor Detection",
    author = "Xu, Yanzhi  and
      Hua, Yueying  and
      Li, Shichen  and
      Wang, Zhongqing",
    editor = "Ku, Lun-Wei  and
      Martins, Andre  and
      Srikumar, Vivek",
    booktitle = "Proceedings of the 62nd Annual Meeting of the Association for Computational Linguistics (Volume 1: Long Papers)",
    month = aug,
    year = "2024",
    address = "Bangkok, Thailand",
    publisher = "Association for Computational Linguistics",
    url = "https://aclanthology.org/2024.acl-long.6/",
    doi = "10.18653/v1/2024.acl-long.6",
    pages = "91--101"
}

@inproceedings{trofi,
    title = "A Clustering Approach for Nearly Unsupervised Recognition of Nonliteral Language",
    author = "Birke, Julia  and
      Sarkar, Anoop",
    editor = "McCarthy, Diana  and
      Wintner, Shuly",
    booktitle = "11th Conference of the {E}uropean Chapter of the Association for Computational Linguistics",
    month = apr,
    year = "2006",
    address = "Trento, Italy",
    publisher = "Association for Computational Linguistics",
    url = "https://aclanthology.org/E06-1042/",
    pages = "329--336"
}

@inproceedings{mohx,
    title = "Metaphor as a Medium for Emotion: An Empirical Study",
    author = "Mohammad, Saif  and
      Shutova, Ekaterina  and
      Turney, Peter",
    editor = "Gardent, Claire  and
      Bernardi, Raffaella  and
      Titov, Ivan",
    booktitle = "Proceedings of the Fifth Joint Conference on Lexical and Computational Semantics",
    month = aug,
    year = "2016",
    address = "Berlin, Germany",
    publisher = "Association for Computational Linguistics",
    url = "https://aclanthology.org/S16-2003/",
    doi = "10.18653/v1/S16-2003",
    pages = "23--33"
}

@inproceedings{bert,
    title = "{BERT}: Pre-training of Deep Bidirectional Transformers for Language Understanding",
    author = "Devlin, Jacob  and
      Chang, Ming-Wei  and
      Lee, Kenton  and
      Toutanova, Kristina",
    editor = "Burstein, Jill  and
      Doran, Christy  and
      Solorio, Thamar",
    booktitle = "Proceedings of the 2019 Conference of the North {A}merican Chapter of the Association for Computational Linguistics: Human Language Technologies, Volume 1 (Long and Short Papers)",
    month = jun,
    year = "2019",
    address = "Minneapolis, Minnesota",
    publisher = "Association for Computational Linguistics",
    url = "https://aclanthology.org/N19-1423/",
    doi = "10.18653/v1/N19-1423",
    pages = "4171--4186"
}

@inproceedings{transformer,
 author = {Vaswani, Ashish and Shazeer, Noam and Parmar, Niki and Uszkoreit, Jakob and Jones, Llion and Gomez, Aidan N and Kaiser, \L ukasz and Polosukhin, Illia},
 booktitle = {Advances in Neural Information Processing Systems},
 editor = {I. Guyon and U. Von Luxburg and S. Bengio and H. Wallach and R. Fergus and S. Vishwanathan and R. Garnett},
 pages = {},
 publisher = {Curran Associates, Inc.},
 title = {Attention is All you Need},
 url = {https://proceedings.neurips.cc/paper_files/paper/2017/file/3f5ee243547dee91fbd053c1c4a845aa-Paper.pdf},
 volume = {30},
 year = {2017}
}

@misc{fuoli,
      title={Metaphor identification using large language models: A comparison of RAG, prompt engineering, and fine-tuning}, 
      author={Matteo Fuoli and Weihang Huang and Jeannette Littlemore and Sarah Turner and Ellen Wilding},
      year={2025},
      eprint={2509.24866},
      archivePrefix={arXiv},
      primaryClass={cs.CL},
      url={https://arxiv.org/abs/2509.24866}, 
}

@article{liang,
title={Using GPT-4 for Conventional Metaphor Detection in English News Texts}, volume={14}, url={https://www.clinjournal.org/clinj/article/view/203}, abstractNote={&amp;lt;p&amp;gt;Metaphor detection presents a significant challenge in natural language processing (NLP) due to the intrinsic complexity of metaphors. In this work, we apply a prompting approach to evaluate GPT-4’s performance on the conventional metaphor identification task. We specifically investigate the effects of prompt variation, output stability, and the role of n-shot prompting. The results indicate that GPT-4’s performance on the metaphor identification task is consistently low across all tested settings, significantly lagging behind the top-performing BERT model. Based on our findings and error analysis, we propose possible approaches for utilizing LLMs and AI assistants&amp;lt;br&amp;gt;in metaphor detection and analysis.&amp;lt;/p&amp;gt;}, journal={Computational Linguistics in the Netherlands Journal}, author={Liang, Jiahui and Dorst, Aletta G. and Prokic, Jelena and Raaijmakers, Stephan}, year={2025}, month={Jul.}, pages={307–341} }

@article{spv1975,
title = {A preferential, pattern-seeking, Semantics for natural language inference},
journal = {Artificial Intelligence},
volume = {6},
number = {1},
pages = {53-74},
year = {1975},
issn = {0004-3702},
doi = {https://doi.org/10.1016/0004-3702(75)90016-8},
url = {https://www.sciencedirect.com/science/article/pii/0004370275900168},
author = {Yorick Wilks}
}

@misc{roberta,
      title={RoBERTa: A Robustly Optimized BERT Pretraining Approach}, 
      author={Yinhan Liu and Myle Ott and Naman Goyal and Jingfei Du and Mandar Joshi and Danqi Chen and Omer Levy and Mike Lewis and Luke Zettlemoyer and Veselin Stoyanov},
      year={2019},
      eprint={1907.11692},
      archivePrefix={arXiv},
      primaryClass={cs.CL},
      url={https://arxiv.org/abs/1907.11692}, 
}

@software{wordfreq,
  author       = {Robyn Speer},
  title        = {rspeer/wordfreq: v3.0},
  month        = sep,
  year         = 2022,
  publisher    = {Zenodo},
  version      = {v3.0.2},
  doi          = {10.5281/zenodo.7199437},
  url          = {https://doi.org/10.5281/zenodo.7199437}
}

@inproceedings{roppt,
    title = "Metaphor Detection with Effective Context Denoising",
    author = "Wang, Shun  and
      Li, Yucheng  and
      Lin, Chenghua  and
      Barrault, Loic  and
      Guerin, Frank",
    editor = "Vlachos, Andreas  and
      Augenstein, Isabelle",
    booktitle = "Proceedings of the 17th Conference of the European Chapter of the Association for Computational Linguistics",
    month = may,
    year = "2023",
    address = "Dubrovnik, Croatia",
    publisher = "Association for Computational Linguistics",
    url = "https://aclanthology.org/2023.eacl-main.102/",
    doi = "10.18653/v1/2023.eacl-main.102",
    pages = "1404--1409"
}

@inproceedings{illinimet,
    title = "{I}llini{M}et: {I}llinois System for Metaphor Detection with Contextual and Linguistic Information",
    author = "Gong, Hongyu  and
      Gupta, Kshitij  and
      Jain, Akriti  and
      Bhat, Suma",
    editor = "Klebanov, Beata Beigman  and
      Shutova, Ekaterina  and
      Lichtenstein, Patricia  and
      Muresan, Smaranda  and
      Wee, Chee  and
      Feldman, Anna  and
      Ghosh, Debanjan",
    booktitle = "Proceedings of the Second Workshop on Figurative Language Processing",
    month = jul,
    year = "2020",
    address = "Online",
    publisher = "Association for Computational Linguistics",
    url = "https://aclanthology.org/2020.figlang-1.21/",
    doi = "10.18653/v1/2020.figlang-1.21",
    pages = "146--153"
}

@book{lakoff1980metaphors,
  author = {Lakoff, George and Johnson, Mark},
  publisher = {University of Chicago press},
  title = {Metaphors we live by},
  year = 1980
}
\clearpage
\appendix

\end{document}